# Deep learning-driven pulmonary artery and vein segmentation reveals demography-associated vasculature anatomical differences

**Authors:** Yuetan Chu[1,2]†, Gongning Luo[1,2]†*, Longxi Zhou[1,2], Shaodong Cao[7], Guolin Ma[8], Xianglin Meng[3], Juexiao Zhou[1,2], Changchun Yang[1,2], Dexuan Xie[5], Dan Mu[9], Ricardo Henao[2], Gianluca Setti[2*], Xigang Xiao[5]*, Lianming Wu[6]*, Zhaowen Qiu[4]*, Xin Gao[1,2]*

**Affiliations:**

[1] Center of Excellence for Smart Health, King Abdullah University of Science and Technology (KAUST), Thuwal 23955-6900, Kingdom of Saudi Arabia.

[2] Computer Science Program, Computer, Electrical and Mathematical Sciences and Engineering Division, King Abdullah University of Science and Technology (KAUST), Thuwal 23955-6900, Kingdom of Saudi Arabia.

[3] Department of Critical Care Medicine, The First Affiliated Hospital of Harbin Medical University, Harbin, China.

[4] College of Computer and Control Engineering, Northeast Forestry University, Harbin, China.

[5] Department of Computer Tomography, The First Affiliated Hospital of Harbin Medical University, Harbin, China.

[6] Department of Radiology, Renji Hospital, School of Medicine, Shanghai Jiao Tong University, Shanghai, China.

[7] Department of Radiology, The Fourth Hospital of Harbin Medical University, Harbin, China.

[8] Department of Radiology, China-Japan Friendship Hospital, Beijing, China.

[9] Department of Radiology, Affiliated Nanjing Drum Tower Hospital of Nanjing University Medical School, Nanjing 210008, China.

* Corresponding author(s). E-mail(s): xin.gao@kaust.edu.sa; gongning.luo@kaust.edu.sa; qiuzw@nefu.edu.cn; wlmssmu@126.com; xxgct_417@126.com; gianluca.setti@kaust.edu.sa

† These authors contributed equally to this work.

**Abstract:** Pulmonary artery-vein segmentation is crucial for disease diagnosis and surgical planning, and is traditionally achieved by Computed Tomography Pulmonary Angiography (CTPA). However, concerns regarding adverse health effects from contrast agents used in CTPA have constrained its clinical utility. In contrast, identifying arteries and veins using non-contrast CT, a conventional and low-cost clinical examination routine, has long been considered impossible. Here we propose a High-abundant Pulmonary Artery-vein Segmentation (HiPaS) framework achieving accurate artery-vein segmentation on both non-contrast CT and CTPA across various spatial resolutions. HiPaS first performs spatial normalization on raw CT volumes via a super-resolution module, and then iteratively achieves segmentation results at different branch levels by utilizing the lower-level vessel segmentation as a prior for higher-level vessel segmentation. We trained and validated HiPaS on our established multi-centric dataset comprising 1,073 CT volumes with meticulous manual annotations. Both quantitative experiments and clinical evaluation demonstrated the superior performance of HiPaS, achieving an average dice score of 91.8% and a sensitivity of 98.0%. Further experiments demonstrated the non-inferiority of HiPaS segmentation on non-contrast CT compared to segmentation on CTPA. Employing HiPaS, we have conducted an anatomical study of pulmonary vasculature on 11,784 participants in China (six sites), discovering a new association of pulmonary vessel anatomy with sex, age, and disease states: vessel abundance suggests a significantly higher association with females than males with slightly decreasing with age, and is also influenced by certain diseases, under the controlling of lung volumes ($p < 0.0001$). HiPaS realizing accurate artery-vein segmentation delineates a promising avenue for clinical diagnosis and understanding pulmonary anatomy in a non-invasive manner.

## Introduction:

Pulmonary and cardiovascular diseases can cause substantial impairment to respiratory function and a substantially increased workload on the cardiovascular[1],[2], representing great threats to human health and well-being globally[3]. Artery-vein segmentation is critical for the diagnosis and treatment of lung diseases[4,5]. It has emerged as an important diagnostic indication for pulmonary and cardiovascular diseases such as pulmonary embolisms and pulmonary arterial hypertension (PAH), and serves as an indispensable anatomical landmark guiding surgical interventions[6]. Beyond disease detection, pulmonary arteries and veins also play a critical role in various medical image analysis tasks, including lesion segmentation and image registration[7-9]. Consequently, considerable endeavors have been devoted to attaining satisfactory artery-vein segmentation. Initially, the Computer Tomography Pulmonary Angiography (CTPA) technique[10] reliant on contrast agents was proposed and subsequently established as the standard approach for artery-vein identification over a long period of time[11]. However, the utilization of contrast agents and arterial cannulations during CTPA can impose substantial metabolic loads on the kidneys and trigger adverse health effects[12],[13]. Furthermore, CTPA remains unsuitable for pregnant women and individuals with an allergy to contrast agents or severe kidney diseases[14]. As a result, an accurate and more universally applicable pulmonary artery-vein segmentation methodology is urgently needed for clinical applications.

Non-contrast computed tomography (CT) has emerged as a widely utilized imaging modality across clinical practice given its affordability and capacity to assist diagnosis[15]. Compared with CTPA, non-contrast CT is faster and more applicable without risks of adverse effects from the contrast agents[16]. While distinguishing pulmonary arteries and veins from non-contrast CT is challenging even for experienced radiologists, recent artificial intelligence (AI) progresses have indicated feasibility in tackling such tasks[17-19]. However, previous approaches show unsatisfactory segmentation accuracy for both vessel trunks and intrapulmonary vessel branches[20]. This can be attributed to the complex topological structures and low radiographic contrast[18]. Current segmentation methods fail to meet clinical necessities for the disease diagnosis such as PAH or distal lesions[18,20], which depend on accurate morphological representations of pulmonary arteries and veins[9, 21, 22]. Alongside diagnostic needs, rising concerns over radiation exposure have precipitated the utilization of low-dose, low-resolution CT (LD-LRCT)[22,24]. However, such LD-LRCT can introduce substantial image noise and the resolution decrease; the former will impede both manual and automated identification of arteries and veins[25-27], while the latter can lead to spatial anisotropy that will greatly degrade the 3D segmentation performances[28]. Therefore, there is a pressing need to develop a novel segmentation framework that can attain anatomically precise and high-abundant segmentation results for pulmonary arteries and veins from varied CT protocols, including non-contrast, CTPA and LD-LRCT scanning.

In this study, we present a novel High-abundant Pulmonary artery-vein Segmentation (HiPaS) framework to achieve rapid and precise artery-vein segmentation results on both non-contrast chest CT and CTPA across various spatial resolutions (Fig 1). This enables contrast-agent-free pulmonary diagnosis, allowing faster, lower-cost examination without the risk of adverse effects from the contrast agents. HiPaS was first self-supervised pre-trained on public chest CT (n=17,817) using a masked autoencoder strategy (Methods) and subsequently trained and externally tested on our multi-center dataset comprising 1,073 CT volumes (315 for CTPA and 758 for non-contrast CT) with meticulous manual artery-vein annotation (875 for training and validation, and 198 for testing). Experiments confirmed the superiority of HiPaS, outperforming other state-of-the-art (SOTA) methods by about 7% and 13% in dice similarity coefficient (DSC), and 15% and 20% in sensitivity for normal and low-resolution CT, respectively. HiPaS also demonstrated exceptional segmentation abundance, detecting about 40% more skeleton length and 100% more vessel branches compared to state-of-the-art methods. We then evaluated HiPaS on external 14 paired CTPA and non-contrast CT (digital subtraction CT pulmonary angiography, DSCTPA) (Methods)[29], showing non-inferior performance achieved by HiPaS on non-contrast CT compared to CTPA.

HiPaS achieving highly accurate and abundant segmentation results enables non-invasive studies of pulmonary vasculature anatomy. Employing HiPaS, we conducted a large-population based anatomical study of pulmonary blood vessels (n=11,784, six sites) in China, uncovering novel associations between vessel abundance and demographic characteristics including sex, age, and disease states. While males show a generally higher vessel abundance, females can exhibit a stronger association with vessel abundance when lung volume is controlled. Vessel abundance also presents a slightly negative association with age, and is impacted by disease states as well. For instance, cardiomegaly shows a positive association with vessel abundance, whereas PAH and chronic obstructive pulmonary disease (COPD) could suggest a negative one. HiPaS not only serves as a robust automatic contrast-agent-free artery-vein segmentation tool for streamlining clinical decisions and mitigating contrast agent hazards but also showcases a new research avenue regarding pulmonary anatomy in a non-invasive modality.

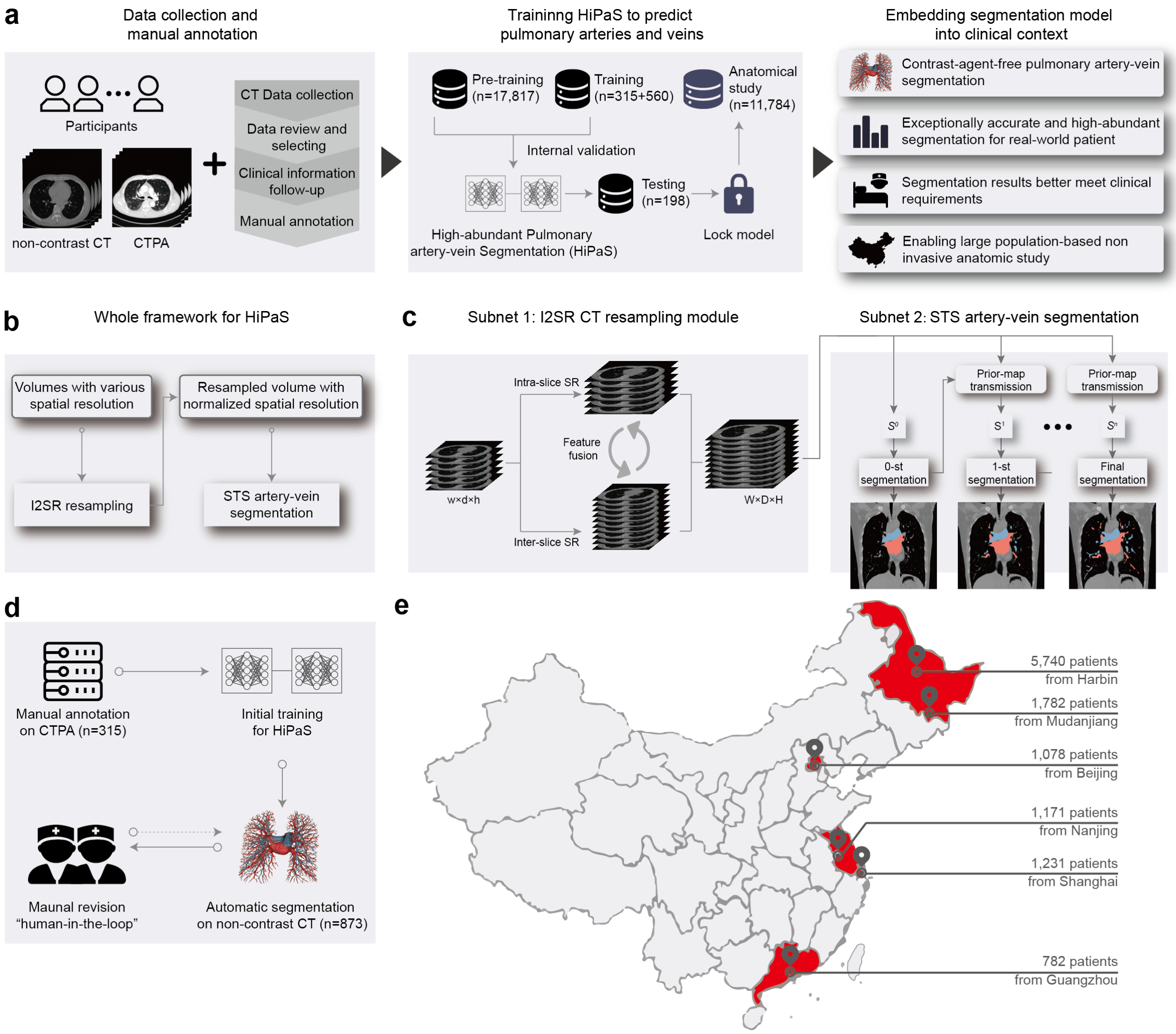

**Figure 1. Illustration of the overall study design. a** Schematic plot for developing and evaluating HiPaS. HiPaS is designed for whole pulmonary artery-vein segmentation on both non-contrast CT and CTPA. Initially, HiPaS was pretrained using 17,817 public CT volumes and subsequently trained on 875 CT volumes, comprising 315 CTPA and 560 non-contrast CT. We utilized 198 external CT volumes for model testing. HiPaS was finally deployed on a clinical cohort of 11,784 patients for anatomical studies. **b** Overall framework of artery-vein segmentation with HiPaS. The HiPaS framework begins with the Inter-and-Intra Slice Super-Resolution (I2SR) module, which resamples CT scans into a normalized space. Following this, the Saliency-Transmission Segmentation (STS) module is used to achieve precise artery-vein segmentation. **c** General structures of the I2SR module and the STS module. **d** Data annotation process. We employed a "human-in-the-loop" strategy for data annotation. Initially, annotations were created for CTPA. HiPaS was trained on these annotations and then deployed on non-contrast CT to generate initial artery-vein segmentation results. Radiologists then reviewed and revised these results to obtain the final segmentation annotations. **e** Datasets for pulmonary anatomical study. We collected 11,784 CT volumes from six different cities in China as a multi-center study.

## Results

### *HiPaS framework and datasets*

Here we present HiPaS for rapid and precise artery-vein segmentation on both non-contrast CT and CTPA (Fig 1.b). HiPaS incorporates a novel Inter-and-Intra-slice Super Resolution (I2SR) module (Supplementary Note 1, 2, and

Supplementary Figure 5.a), and a Saliency-Transmission Segmentation (STS) module (Supplementary Note 1, Supplementary Figure 5.b, c) that are jointly employed to produce accurate segmentation. To resolve spatial anisotropy due to varying inter- and intra-resolution unbalance, the I2SR module first reconstructs the original CT scans into a space with normalized spatial resolution. Unlike traditional reconstruction approaches, the I2SR module can mutually learn the representation of inter-slice and intra-slice multi-scale features to improve the accuracy and spatial consistency in the reconstruction process. We additionally introduce an image enhancement block to increase the reconstruction fidelity of vascular structures and robustness to image noise. In the artery-vein segmentation process, to address previous limitations related to long-range context correlation and diverse topological structures, inspired by manual labeling, which progresses from low-level to high-level vessels, we design an innovative cascading multi-stage saliency-transmission segmentation strategy. Initially, the vascular tree is divided into distinct levels of vessel branches (Methods), and sequentially assign each branch level as the segmentation target for each respective segmentation stage. The segmentation probability map from lower-level vessels is used as a prior for higher-level vessel segmentation by integrating the probability map with the CT volumes as the input for subsequent networks. This approach allows us to progressively achieve accurate segmentation results.

Given the difficulties in identifying arteries and veins from non-contrast CT scans, precise manual annotation is laborious, expensive, and challenging[18]. It demands substantial endeavor expendable only by expert radiologists, with an estimated annotation time of 4 to 5 hours per case. To our knowledge, no publicly available large-scale dataset currently exists about whole pulmonary artery-vein segmentation for both non-contrast CT and CTPA modalities. Here, for the first time, we establish a multi-center CT dataset with high-abundant pulmonary artery-vein annotations (Table 1), which includes almost all the visible arteries and veins on the chest CT. The CT volumes employed in this study were obtained from three different sites and were acquired from September 2018 to August 2023. A selected group of experienced thoracic radiologists first undertook manual annotation of the arteries and veins for the CTPA subset (n = 315). Our algorithm was first trained and validated utilizing these annotated CTPA to get an initial model (Methods). Subsequently, the initially trained model was transferred and deployed on the remaining larger corpus of non-contrast CT (n = 758) (the model transferring is described in the Method section) to generate initial segmentation results. The radiologists then reviewed and corrected these initial results to obtain the final high-abundant artery-vein segmentation for a total of 1,073 CT volumes (about half an hour on average for one case labeling) (Fig. 1.d, Supplementary Figure 8).

A multi-step training strategy was implemented to optimize HiPaS. Pre-training on extensive public chest CT (n = 17,817) with the masked auto-encoder approach (Methods) was initially employed to increase the generalizability of HiPaS to real-world data. Subsequently, to improve the anatomical perception ability of HiPaS for arteries and veins, HiPaS was first trained on CTPA scans, followed by fine-tuning on non-contrast CT scans (Fig. 1.a) using the Harbin and Mudanjiang cohort (Table 1). HiPaS testing was performed on the external Guangzhou cohort (n=198). Quantitative evaluation of HiPaS included several indices: dice similarity coefficient (DSC) for whole arteries and veins, as well as for intrapulmonary arteries and veins, respectively; sensitivity, 1 – specify (false positive ratio), artery-vein misclassification (MCS) ratio; detected proportion of branch counts (BC) and vessel skeleton length (SL)[30]; and 95% Hausdorff Distance (HD95)[31] (Supplementary Table 5). Details regarding the implementation of evaluation indices are described in the Methods section.

| | Data source | Number | Protocol | Age (y) | Male/Female (%) | Vendor | Utilization |
|---|---|---|---|---|---|---|---|
| Public chest CT dataset (n=17,817) | RAD chest dataset | 13,000 | - | - | - | - | Model pretraining |
| | RSNA PE CT dataset | 4,817 | - | - | - | - | Model pretraining |
| Established dataset with annotations (n=1,073) | Harbin | 315 | CTPA | 54.3±7.9 | 37.1/62.9 | SIEMENS | Model training |
| | Harbin | 343 | NCCT | 53.9±7.6 | 51.9/48.1 | SIEMENS | Model training |
| | Mudanjiang | 217 | NCCT | 56.5±7.2 | 49.6/50.4 | SIEMENS | Model training |
| | Guangzhou | 198 | NCCT | 60.0±9.3 | 36.0/64.0 | Philips, GE | Model testing |
| Clinical association cohort (n=11,784) | Harbin | 5,740 | CTPA&NCCT | 51.8±6.1 | 41.4/58.6 | SIEMENS, UIH | Anatomical study |
| | Mudanjiang | 1,782 | NCCT | 52.7±5.8 | 55.1/44.9 | SIEMENS, Philips | Anatomical study |
| | Guangzhou | 782 | CTPA&NCCT | 57.9±8.6 | 34.4/65.6 | Philips, GE | Anatomical study |
| | Beijing | 1,078 | CTPA&NCCT | 54.3±7.9 | 47.4/52.6 | SIEMENS | Anatomical study |
| | Shanghai | 1,231 | CTPA&NCCT | 51.3±9.5 | 46.2/53.8 | TOSHIBA, UHI | Anatomical study |
| | Nanjing | 1,171 | CTPA&NCCT | 51.6±11.4 | 62.5/37.5 | Philips, UIH | Anatomical study |

**Table 1. Detailed information of datasets for model establishing and anatomical study.** "NCCT" = non-contrast CT. Patient age is summarized as $\text{mean} \pm \text{std}$.

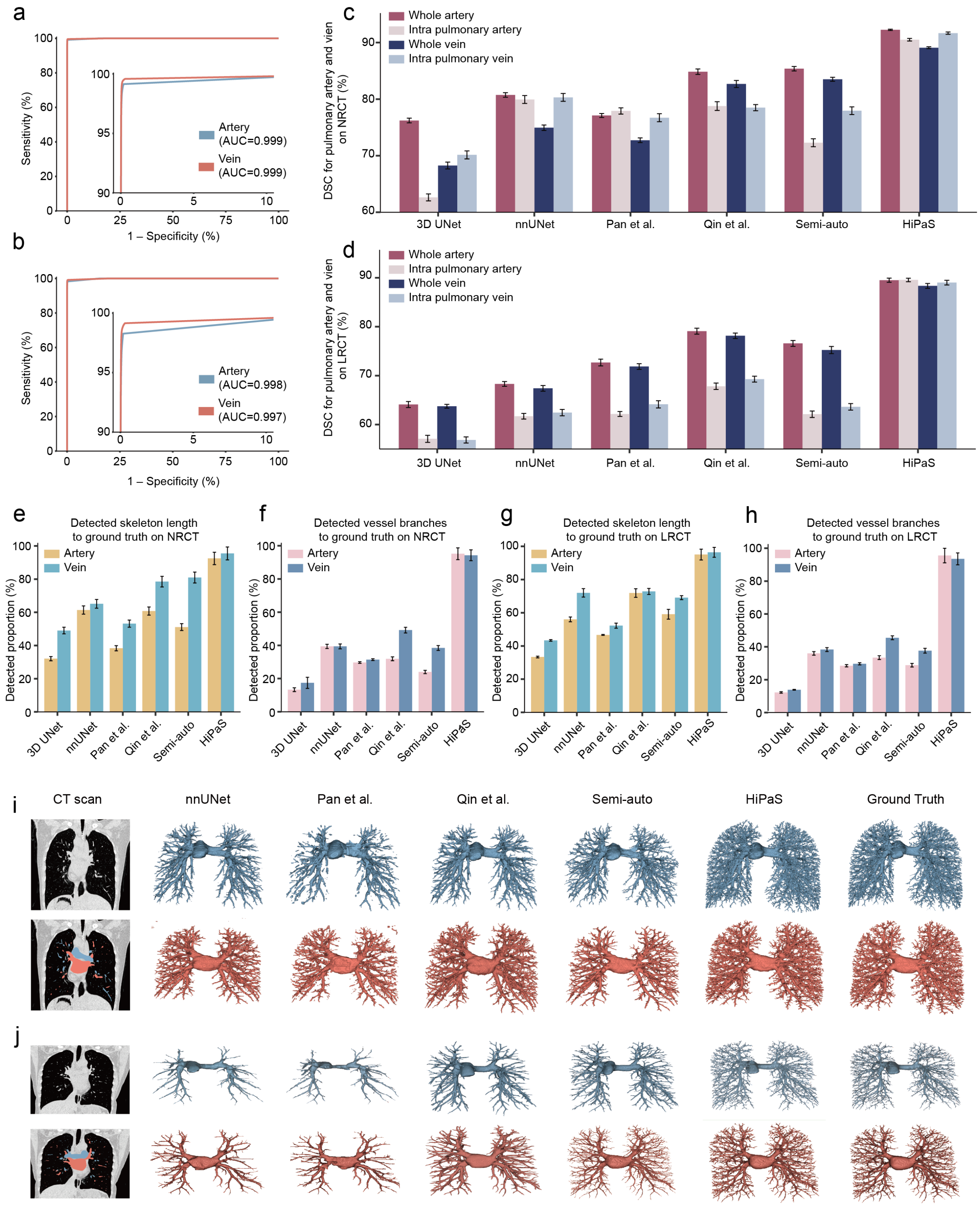

**Figure 2. External testing on the non-contrast CT of the Guangzhou cohort. a, b** Receiver operating characteristic (ROC) curves of pulmonary arteries and veins on normal-resolution CT (NRCT) (**a**) and low-resolution CT (LRCT) (**b**). We zoom in on the ROC curve near the top left corner for better visualization. **c, d** Performance comparison with existing methods in terms of dice similarity coefficient (DSC) for whole arteries and veins, as well as intrapulmonary arteries and veins on both NRCT and LRCT. The error bars here and below indicate 95% Confidence Interval (CI). **e, f** Comparison with existing methods in terms of detected proportion of skeleton length and branch counts for arteries

and veins on NRCT. **g, h** Comparison with existing methods in terms of detected ratio of skeleton length and branch counts for arteries and veins on LRCT. **i, j** The CT scans and illustrations merged with artery-vein segmentation. Arteries are marked with blue color and veins are marked with red. We also show the 3D rendering of the arteries (first row) and veins (second row) segmentation results on NRCT (**i**) and LRCT (**j**) achieved by different methods. HiPaS can achieve more consistent results with ground truth with more abundant vessel branches.

***Performance evaluation on external dataset***

We present both quantitative (Fig 2.a-h, Supplementary Data Table 5) and qualitative (Fig 2.i, j, Supplementary Figure 7) results of HiPaS on the non-contrast CT segmentation against state-of-the-art methods. The compared approaches include the 3D UNet[32], nnUNet[33], and the algorithms introduced by Pan et al.[19] and Qin et al.[18], and the semi-automatic (Semi-auto) achieved by radiologists (Method). All segmentation approaches except the semi-automatic segmentation underwent initial training and optimization on the same training datasets, followed by testing on the same external Guangzhou cohort. The testing cohort was further separated into two sets according to the spatial resolution and scanning doses: the first set (n = 142) contained CT images acquired under normal-dose (scanning voltage ⩾ 120kV) and normal resolution (inter-slice thickness = 1.00mm) CT scans (NRCT), while the second set (n = 56) contained relatively lower-dose (scanning voltage = 100kV) and lower-resolution (inter-slice thickness ⩾ 1.25mm) CT scans (LRCT). To ensure unbiased assessment, testing cases were strictly excluded from all training and fine-tuning procedures.

Our method achieves the best performance under most metrics (Supplementary Table 5, first table) on the NRCT set, with a mean DSC of 92.25% (95% CI 92.12%-92.38%) / 89.09% (95% CI 88.91%-89.27%) (the results here and below are presented in artery/vein terms, respectively) (Fig 2.c), and the area under curve (AUC) score is 99.93% / 99.91% (Fig 2.a). Compared to other methods, our approach achieves significantly higher accuracy and sensitivity ($p < 0.0001$) and can detect much more vessel skeleton lengths and branches (Fig 1.e, f). Specifically, compared to best available SOTA method, HiPaS can detect about 25% / 56% relatively more skeleton length and 86% / 130% relatively more vessel branches, with a detected ratio of 95.20% (95% CI 91.35%-99.05%) / 95.11% (95% CI 91.26%-98.95%) for vessel skeleton length and 94.28% (95% CI 91.04%-97.53%) / 95.21% (95% CI 91.65%-98.76%) for branches (Fig 2.e, f). These results collectively demonstrate that our segmentation results are very close to the ground truth, which can reflect the realistic anatomical structures.

In terms of robustness, HiPaS is validated on LRCT scans (Supplementary Data Table 5, second table). Remarkably, our technique yields superior performance even under the challenging condition, attaining a mean DSC of 89.51% (95% CI 88.79%-90.23%) / 88.34% (95% CI 87.49%-89.19%) (Fig 1.d), which is comparable to the segmentation performance on normal-resolution CT scans ($p = 0.103$). Compared to other methods, our approach exhibits significantly better results ($p < 0.0001$) and can detect the most vessel branches, where the AUC score is 99.82% / 99.74% (Fig 2.b), the sensitivity reaches a score of 97.24% (95% CI 97.10%-97.38%). Compared to the best available state-of-the-art method, HiPaS can detect about 40% / 42% relatively more skeleton length and 106% / 160% relatively more vessel branches, with a detected ratio of 96.39% (95% CI 93.40%-99.39%) / 95.06% (95% CI 91.79%-98.33%) for vessel skeleton length and 93.51% (95% CI 89.88%-97.14%) / 95.36% (95% CI 90.92%-99.80%) for branches (Fig 2.g, h). This proves the robustness of HiPaS for low-resolution scanning, suggesting the potential applicability in scenarios where scanning condition is sub-optimal.

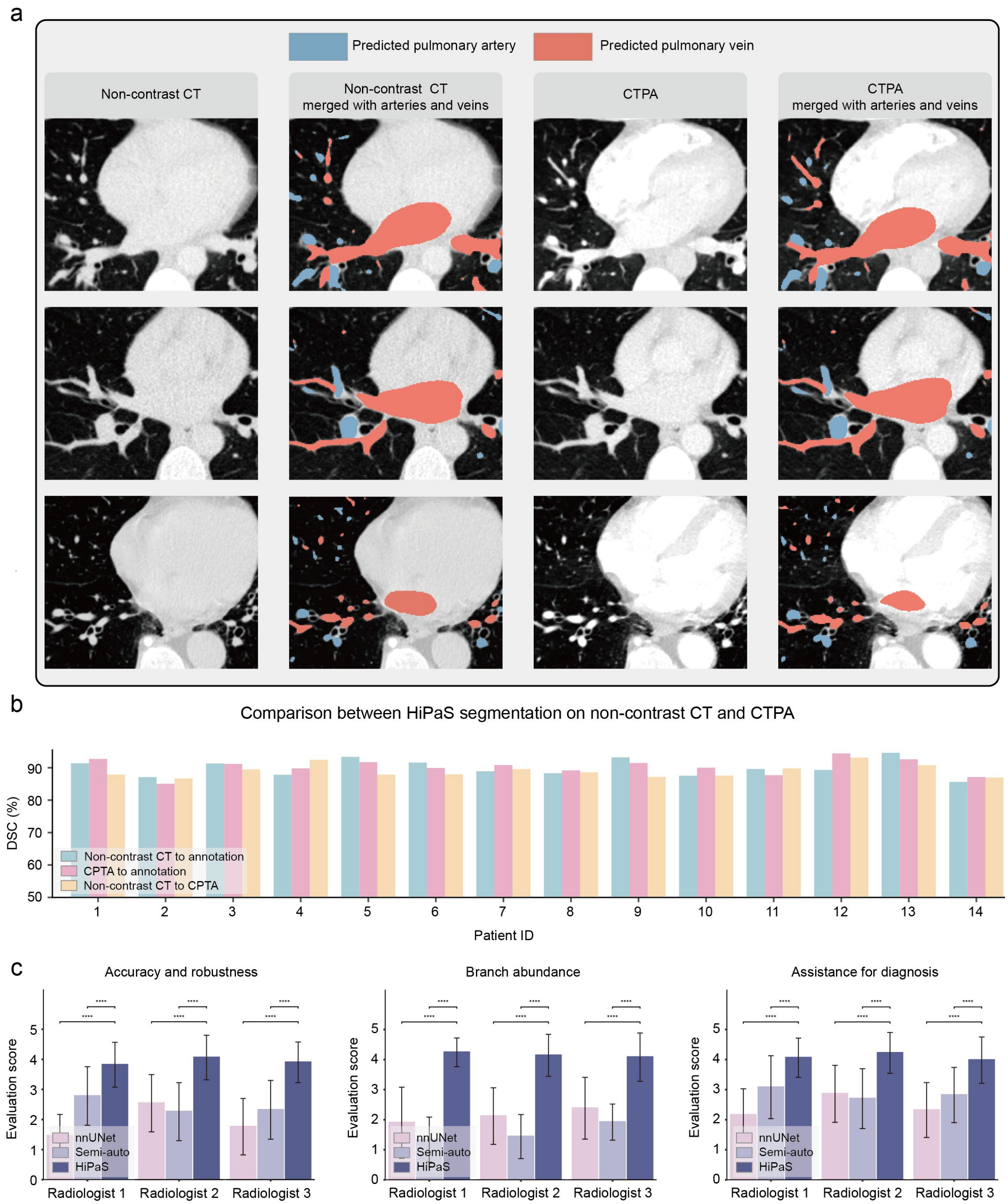

**Figure 3. Segmentation comparison against CTPA, and clinical evaluation for HiPaS. a** Comparison of artery-vein segmentation on paired non-contrast CT and CTPA achieved from DSCTPA. HiPaS can identify arteries and veins directly from non-contrast CT, whose performance is non-inferior to the segmentation on CTPA. **b** Quantitative comparison of segmentation results between non-contrast CT and paired CTPA from the same patient. Dice similarity coefficient (DSC) is calculated in three scenarios: 1) between segmentation on non-contrast CT and the corresponding annotations; 2) between segmentation on CTPA and the corresponding annotations; and 3) between segmentation from non-contrast CT and CTPA. **c** Clinical evaluation of HiPaS. Three radiologists from distinct hospitals independently assessed the segmentation results derived from the three methods, nnUNet, semi-automatic segmentation, and HiPaS.

The specific method corresponding to the segmentation results remained undisclosed to the radiologists, ensuring unbiased evaluations. The assessment encompassed three key indicators: segmentation accuracy and robustness, vessel branch abundances, and diagnostic assistance. Mann-Whitney U tests were done between each method. *P*-values are specified as *$p < 0.05$, ** $p < 0.01$, ***$p < 0.001$, **** $p < 0.0001$, NS, not significant.

***Performance comparison between non-contrast CT and CTPA***

To rigorously compare the segmentation performance of HiPaS on non-contrast CT versus CTPA, the clinical gold standard, we conducted a quantitative analysis of the segmentation results from paired and aligned non-contrast CT and CTPA scans. These scans were obtained from digital subtraction CTPA (DSCTPA)[29] performed on external fourteen patients (Methods). Radiologists manually annotated arteries and veins on CTPA and non-contrast CT scans. Several metrics were calculated to evaluate the performance and the differences in segmentation results generated by HiPaS on non-contrast CT versus CTPA: the DSC between the segmentation results achieved by HiPaS and the manual annotations on non-contrast CT, the DSC between the segmentation results achieved by HiPaS and the manual annotations on CTPA, and the DSC between the segmentation results on non-contrast CT and CTPA achieved by HiPaS, respectively.

Qualitative and quantitative results are presented in Fig 3.a, b. Experiments demonstrate that the segmentation results of arteries and veins on non-contrast CT and CTPA are highly consistent. The segmentation from non-contrast CT and CTPA achieves an average DSC of 89.95% (95% CI: 89.28-90.63%) and 90.24% (95% CI: 89.62-90.86%), and their performances are comparable ($p = 0.633$). The segmentation results on non-contrast CT and CTPA also achieve high similarity and consistency, with an average DSC of 88.98% (95% CI: 88.47-89.48%). The experiment indicates the potential of HiPaS with non-contrast CT to serve as a viable alternative to CTPA in segmenting arteries and veinsy.

***Prospective clinical utility***

To assess the clinical value of HiPaS, we invited three radiologists from three leading institutions in China (the First Affiliated Hospital of Harbin Medical University, the Fourth Affiliated Hospital of Harbin Medical University, and Guangdong Provincial People's Hospital) to conduct a prospective evaluation. The study included 50 representative cases from routine clinical practice (Supplementary Table 6). The evaluation was conducted across three objective indicators: accuracy and robustness, branch abundance, and assistance for diagnosis (Supplementary Table 7). A 5-point scale was utilized for scoring, with manual annotation serving as the referenced gold standard. A 5 score indicates excellent segmentation with high consistency with the gold standard, and 0 indicates unacceptable. For each case, we presented the segmentation results achieved by nnUNet, semi-automatic segmentation, and HiPaS to all the radiologists in a blinded fashion to enable unbiased comparative analyses. Fig 3. c presents statistical results from the evaluation. HiPaS demonstrates superior performance across all indicators for all participating radiologists compared to nnUNet and semi-automatic segmentation. Our proposed method achieves the highest accuracy and better facilitates diagnostic tasks as evidenced by the evaluation scores. These results highlight the clinical value of HiPaS.

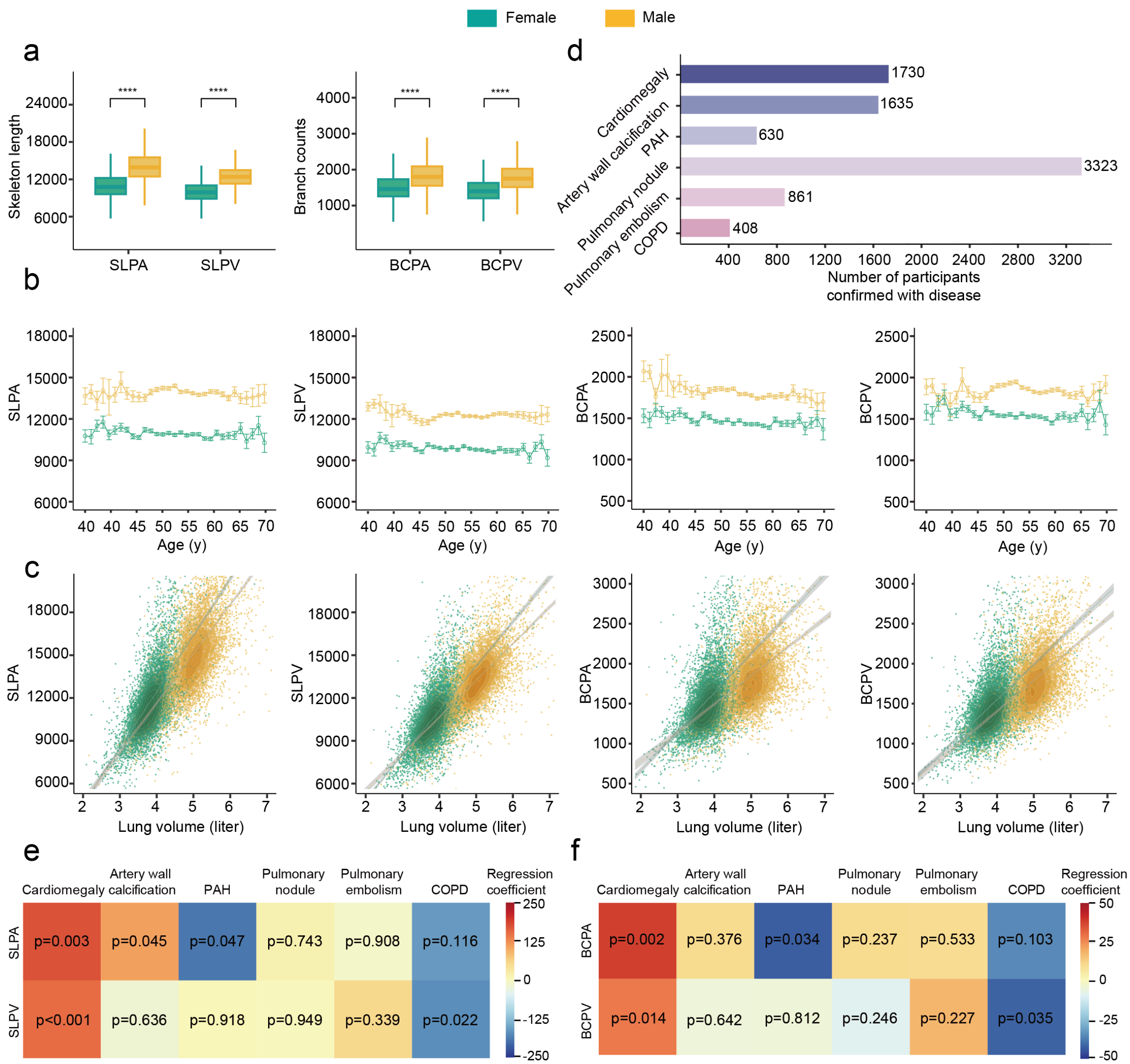

**Figure 4. Association of vessel abundance with sex and age on 11,784 participants.** We include four indices, skeleton length of pulmonary artery (SLPA), skeleton length of pulmonary vein (SLPV), branch count of pulmonary artery (BCPA), and branch count of pulmonary vein (BCPV) to represent the blood vessel abundance, and used the lung volume as the controlling. **a** Boxplot of the distribution of four indices between males and females. **b** Pulmonary vessel abundance across different ages. Error bars show the standard error of mean (SEM). **c** Association and linear regression of vessel abundance compartments with lung volume. **d** Numbers of confirmed disease states for the involved participants. "PAH" = pulmonary artery hypertension, "COPD" = chronic obstructive pulmonary disease. **e, f** Association between vessel abundance (**e** for skeleton length and **f** for branch counts) compartments with disease states. Values of the regression coefficients are indicated by colors. *P* values are specified for each item.

### ***HiPaS enables vasculature anatomical study in large cohorts***

HiPaS achieving sufficiently high-abundant and accurate segmentation results enables a systematic investigation of the anatomy of pulmonary blood vessels. Although the physiology and anatomy of the respiratory system have been studied[33], quantitative assessment of pulmonary vessels across sex, age, and diseases, to our knowledge, has not been reported. Here we conducted a substantial, multi-center anatomical study on a large-scale CT dataset comprising 11,784 participants from six representative sites in China (Harbin, Mudanjiang, Guangzhou, Beijing, Shanghai, and Nanjing) (Fig 1.e, Table 1, Supplementary Figure 9). To ensure standardized analysis, all CT volumes were resampled into the normalized space (Methods), after which HiPaS was implemented to obtain the artery-vein segmentation of these volumes. Quantification of pulmonary vascular abundance was evaluated using four statistical indices: skeleton length of pulmonary artery (SLPA), skeleton length of pulmonary vein (SLPV), branch count of pulmonary artery (BCPA), and branch count of pulmonary vein (BCPV), which were calculated from the segmentation results. Quantitative comparisons and statistical analyses of the association between pulmonary vessel abundance and lung volumes, sex, age, and diseases were carried out utilizing these four defined indices. Here lung volume served as the control for the correlation study, defined as the volume of segmented lung regions on CT scans (Methods). Sex was coded as a binary variable (male = 1, female = 0) for the correlation calculations. We included six common pulmonary and cardiovascular diseases (cardiomegaly, artery wall calcification, PAH, pulmonary nodule, pulmonary embolism, and COPD) (Fig 4.d). We also performed binary-encoding for each disease state respectively (with disease = 1, without disease = 0). The statistical results for the whole dataset are presented in Fig 4 and Supplementary Table 8, and the results for the 1,073 CT volumes with manual annotations are presented in Supplementary Figure 10 and Supplementary Table 9 as a reference.

Regarding the entire study population, pulmonary vessel abundance exhibited a strong correlation with lung volume for both males and females ($p < 0.0001$). Males have significantly longer vessel-skeleton lengths (13985±2794 / 12257±2407 for males and 10977±2407 / 9871±1916 for females, respectively) ($p < 0.0001$) and more vessel branches (1862±472 / 1805±462 for males and 1552±470 / 1468±430 for females) ($p < 0.0001$) (Fig 4.a), while females exhibited a larger regression coefficient of skeleton length and branch counts with lung volume compared to males (3368±45 / 2652±36 of vessel skeleton length cm per liter for females and 3004±50 / 2222±34 cm per liter for males, and 485±10 / 437±10 of branch counts per liter for females and 362±10 / 397±9 per liter for males, respectively; $p < 0.0001$) (the correlation coefficient for sex is -838±55 / -565±40 for vessel skeleton length and -195±12 / -160±11 for branch counts, when encoding male as 1 and female as 0). These results indicated that given the same lung volume, females could exhibit greater vessel abundance, including longer vessel skeleton length and more branch counts, compared to males. Pulmonary vessel abundance also exhibited strong correlations with age in both males and females (Fig 4.b, and Supplementary Table 8). As age increases, pulmonary vessel abundance, as quantified by all four indices, showed a slightly declining trend ($p < 0.0001$). Benefiting from HiPaS, aging effects on the vascular system[34] are proven in radiomics.

Pulmonary vessel anatomy also demonstrates significant associations with certain diseases (Fig. 4.e, f). For instance, cardiomegaly and artery wall calcification show a positive correlation with overall vessel abundance (172 ± 58/150 ± 42 for skeleton length, and 38 ± 12/28 ± 11 for branch counts) and artery abundance (106 ± 53 for artery skeleton length), respectively. PAH is associated with a significant decrease in artery abundance (-192 ± 96 for skeleton length and -43 ± 20 for branch counts), while COPD negatively correlates with vein abundance (-170 ± 74 for vein skeleton length and -42 ± 20 for branch counts). This negative association implies that these two diseases might contribute to the remodeling and removal of distal pulmonary vessels[37] [38]. Furthermore, diseases such as pulmonary nodules and pulmonary embolism exhibit less significant associations with vessel abundance, indicating they may not solely substantially affect the anatomy of distal vessels. HiPaS facilitates quantitative monitoring of the changes in pulmonary vasculature anatomy via non-contrast CT, which can help in understanding the disease's influence on blood vessels. Many studies also demonstrate that these changes can prove valuable for treatment monitoring, prognosis, and therapeutic strategies targeting pulmonary and cardiovascular diseases[37],[39],[40].

**Discussion**

In this work, we have proposed an innovative framework, HiPaS, designed for the highly accurate and abundant segmentation of pulmonary arteries and veins in both non-contrast CT and CTPA. Pulmonary artery-vein segmentation is crucial for clinical diagnosis and surgical planning, but has traditionally relied on CTPA in clinical practice. Due to the low image contrast and complex vascular structures, directly segmenting arteries and veins from non-contrast CT has long been considered infeasible by radiologists and computer-aided diagnosis (CAD) systems. Here we present the feasibility of using HiPaS to segment arteries and veins directly on non-contrast CT, with non-inferiority than segmentation on CTPA. Extensive experiments on external datasets have demonstrated the superior performance of our framework, achieving sufficiently high-abundant and accurate segmentation results, and enabling systematic investigation of the anatomical study of pulmonary blood vessels. By facilitating accurate clinical segmentation without contrast-agent utilization, HiPaS facilitates rapid, accurate, and non-invasive pulmonary disease diagnosis and surgical planning.

The success of HiPaS can be attributed to our framework design and training strategy. The introduction of the I2SR module addresses potential blurring and spatial anisotropy due to low-resolution scanning, while the enhancement block provides robust priors to aid segmentation in noisy data. Inspired by the manual labeling process, we propose the STS module, which progressively achieves the whole pulmonary segmentation by utilizing lower-level vessel segmentation as a priori for higher-level segmentation. This improves the anatomical perception of the network, enabling it to model long-range vascular correlations and morphological differences between vessel trunks and intrapulmonary branches. Furthermore, transfer learning from CTPA to non-contrast CT allows the network to learn morphologies of arteries and veins, facilitating identification and segmentation in low-contrast non-contrast CTs. Additionally, our "human-in-the-loop" strategy greatly reduces manual annotation efforts, (from about four hours per case with fully manual annotation, to about half an hour with the help of "human-in-the-loop"), allowing us to efficiently obtain large-scale, meticulously labeled datasets. This enables large data-driven training of our models to achieve optimal performances.

Furthermore, our research represents an initial step toward understanding the influence of demographic characteristics, including sex, age, and diseases, on pulmonary vessel anatomy. While previous studies report sex differences in pulmonary anatomy, such as larger lung volumes and airway sizes in males compared to females[41], similar studies on vascular differences between sexes are notably lacking. Our anatomical studies can provide some inspirations for this problem. Moreover, our findings on sex differences in pulmonary blood vessels may help understand the disparities in disease occurrence and treatment, which may aid in the development of more personalized diagnostic and therapeutic strategies. Many studies have reported sex differences in disease incidence and treatment for conditions like pulmonary embolism, hypertension, and lung cancer [42-46]. For example, females have a higher PAH incidence compared to same-age males[42]. We hope our findings can help provide a perspective to understand such differences. Additionally, both our clinical experience and previous research have reported that females may experience a higher likelihood of hemorrhage during lung surgery and pulmonary transthoracic biopsy compared to males[47], potentially due to a higher pulmonary vessel density in females. This also underscores the significance of more detailed surgical planning with HiPaS for females to potentially reduce intraoperative hemorrhage. We hope our findings can help understand these sex differences, contributing to the establishment of more targeted, sex-differentiated disease prevention and treatment strategies in clinical practice.

HiPaS also facilitates the observation of disease impact on pulmonary blood vessels using non-contrast CT. Many diseases can cause pulmonary vascular remodeling and change vessel abundance. For example, our study identifies that cardiomegaly and artery wall calcification can show a positive correlation with overall vessel abundance. Cardiomegaly can compress lung volume[48], thus increasing overall vessel density. Meanwhile, cardiomegaly is also strongly linked to declining cardiac function, with increased vessel density potentially serving as a compensatory response[48]. Calcification, besides being compensatory, also enhances radiological signals[49] and thus maybe increases their visibility on CT scans. We also find that PAH and COPD can reduce vessel abundance when lung volumes are controlled, which can be attributed to the small vessel atrophy and removal caused by these two diseases[37] [38]. Quantitative assessment of these changes is crucial for understanding disease extent and severity. For instance, the loss of pulmonary vessels has been shown to be related to the severity of PAH in patients with COPD[50]. Some studies also suggest vessel alterations can precede alveolar destruction, indicating that monitoring pulmonary vessels may provide early insights into COPD progression[39][40]. This vessel assessment also has significant prognostic value. For instance, reduced small vessel volumes in PAH are correlated with worse survival, which can aid risk stratification and management[37]. Traditionally, such quantitative identifications are performed using gadolinium-enhanced MRI or CTPA[37] [38]. HiPaS, however, offers a cost-effective and more adaptable alternative. We anticipate that HiPaS-guided artery-vein segmentation will significantly influence future clinical practices.

There remain opportunities to further improve HiPaS in the future. Although HiPaS is trained and tested on a multi-center dataset, it currently only includes Chinese hospital CT scans. Due to data limitations, scans with clinical information from other countries or continents were unavailable in our study. Validating the performance of the model and anatomical studies on more geographically diverse international cohorts is our future goal. The anatomical study also faces some limitations. While HiPaS can accurately identify small distal pulmonary vessels, it only detects visible vessels in CT and cannot distinguish vessel walls or blood flow. For instance, blood vessel wall thickening is a sign of vascular degeneration[51], but HiPaS cannot identify such changes. More precise techniques, like microscopy, are required to assess these conditions. Additionally, the anatomical study is constrained by CT resolution, which will exclude vessels with diameters below the resolution limit. Super-resolution with deep-learning methods cannot overcome this physical limitation. With advancements in CT scanning technology, we will conduct a second confirmation of our anatomical studies to prove its efficiency.

HiPaS demonstrates the promising potential of using widely available, low-cost, low-risk non-contrast CT for accurate artery-vein segmentation. We hope that HiPaS will assist radiologists with disease diagnosis and surgical planning, while also catalyzing a transformation in clinical artery-vein segmentation from contrast-enhanced CTPA to more convenient non-contrast approaches.

## Methods

### *Problem formalization*

Let $X \in \mathbb{R}^{w \times d \times h}$ be the input CT scans. Our goal is to computationally achieve the artery-vein segmentation results $P$

$$P = Seg_{\mathrm{artery-vein}}(X).$$

Here we propose an integrative workflow to divide the problem into two-mappings: spatial normalization $F: X \to Y$ and segmentation $G: Y \to P$. Here $Y$ is the resampled CT scan, $Y \in \mathbb{R}^{W \times D \times H}$, with the normalized spatial resolution $\frac{334}{512} \times \frac{334}{512} \times 1.00\ \mathrm{mm}^3$. This spatial resolution is sufficiently high to preserve information in most cases and ensures that the resampled space is large enough to fully accommodate most human lungs. Then the entire process can be denoted as

$$X^{w \times d \times h} \xrightarrow{F} Y^{W \times D \times H} \xrightarrow{G} P^{W \times D \times H}.$$

The overview workflow is shown in Fig 1.c. In the inference phase, we first resample the raw input CT to the normalized space using trilinear interpolation. For inter-slice thicknesses greater than 2 mm, we apply I2SR for inter-slice super-resolution instead. STS segmentation is then performed on the resampled normalized data. The predicted segmentations will be resampled back to the original input resolution to compare our segmentation results and the ground truth labeled on raw resolution space.

### *Model pretraining with masked autoencoder*

Self-supervised pre-training on large datasets has demonstrated improved model generalizability and downstream task performance. Here we implement masked autoencoder (MAE), a highly effective pre-training approach[52], on our proposed artery-vein segmentation network with sparse convolution[53]. We mask parts of the input data and then guide the network to learn effective representation by training to predict the masked parts. The sparse convolutional is employed in the encoder to avoid the distribution shift due to irregular image masks[53]. We employ identical hyper-parameter values for mask patch size ($32 \times 32 \times 32$) and masking ratio (60%) as proposed in [53], alongside an L2 objective function to optimize the difference between the reconstructed input and the original unmasked input. We utilize the public chest CT released in [60] (n=13,000) and [61] (n=4,817) (17,817 in total) as the pretraining dataset. When transferring to artery-vein segmentation, we fix the encoder parameters of the pretrained model and solely optimize the decoder.

### *CT spatial-normalization with I2SR module*

In this section, we describe the structures of our proposed I2SR (Fig 1.c, left, and Supplementary Figure 5.a), as a pre-processing step for the subsequent segmentation module. Unlike previous CT super-resolution techniques that treat CT as two-dimensional images or three-dimensional volumes, I2SR fully leverages the context of CT slices and treats CT volumes as a sequence of two-dimensional slices with position information encoded. This approach not only improves the accuracy of the intra-slice reconstruction results but also maintains the consistency and continuity of inter-slice information.

I2SR mainly consists of four parts: feature extraction (FE) block, feature interpolation (FI) block, inter- and intra-feature fusion (I2FF) modules, and final high-resolution reconstruction (RE) block. Given two contiguous CT slices in the low-resolution CT scan, denoted as $\mathrm{slice}_1^{\mathrm{lr}}$ and $\mathrm{slice}_2^{\mathrm{lr}}$, our first step involves utilizing the feature extraction block to encode the input slices into the latent representations. This block architecture consists of residual-connected convolutional layers[54], outputting embedding features of identical dimensions to the input volumes,

$$(L_1, L_2) = \mathrm{FE}(\mathrm{slice}_1^{\mathrm{lr}}, \mathrm{slice}_2^{\mathrm{lr}}).$$

Here $L_1$ and $L_2$ are the latent representations of input slices. We then perform feature interpolation with the FI block, upsampling these latent representations to the target resolution and spatial dimensions, yielding an initialized temporary HR representation $(H_1, \cdots, H_n)$.

$$(H_1, \cdots, H_n) = \mathrm{FI}(L_1, L_2).$$

While the initial HR representations provide a preliminary reconstruction, further optimization is imperative to enhance result fidelity through simultaneous inter-slice and intra-slice contextual learning. We achieve this via iterative interactive I2FF modules, as illustrated in Supplementary Figure 5.a. Each I2FF component comprises parallel 3D convolutional layers with kernel sizes of $k_1 \times k_1 \times 1$ and $1 \times 1 \times k_2$ designed to integrate information across both the spatial and depth dimensions, respectively. The output of two convolutional layers will be concatenated along the channel dimension. We additionally employ two cascaded 3D convolution layers with $3 \times 3 \times 3$ and $1 \times 1 \times 1$ kernels to further fusion features between channels. Layer normalization and ReLU activation functions are also implemented throughout for improved conditioning and non-linearity. The I2FF blocks are iteratively operated on the HR representations to perform feature fusion, and the final HR representations are sent into the RE block to achieve the final results.

$$(\mathrm{slice}_1^{\mathrm{re}}, \cdots, \mathrm{slice}_n^{\mathrm{re}}) = \mathrm{RE} \circ (\mathrm{I2FF})^{\,l}(H_1, \cdots, H_n).$$

Here $(\mathrm{slice}_1^{\mathrm{re}}, \cdots, \mathrm{slice}_n^{\mathrm{re}})$ denote the final HR reconstruction results, and $l$ is the iterative numbers. Supplementary Note 1, Note 2, and Supplementary Figure 5 provide specific network architectures and training strategies employed in our experimental setting. To enhance anatomical fidelity and spatial consistency, we optimize the reconstructed images in both the image domain and vascular-structural space via our image enhancement training strategy described as follows.

### *Image enhancement and objective function for I2SR*

We propose a novel technique integrating image enhancement to improve the reconstruction fidelity of vascular anatomy. The Frangi filter[55] is utilized to extract the edges and vascular structures from the CT scan (Supplementary Figure 6.a), and our object is to keep these extracted features consistent before and after reconstruction, ensuring accurate vascular reconstruction via the I2SR module. Denoting the Frangi filter operator as $\mathrm{O}_F(\cdot)$, the objective function can be formulated as

$$\mathcal{L}^{vessel} = \| O_F(\mathrm{slice}^{\mathbf{re}}) - O_F(\mathrm{slice}^{\mathrm{gt}}) \|^2.$$

Here $\mathrm{slice}^{gt}$ is the referenced high-resolution CTs. However, the Frangi filter includes high-order complex operations including Hessian eigenvalue calculation, making back-propagation difficult in the training process[56]. In our practice, we develop a convolution-based operator comprising multiple lightweight $3 \times 3$ convolutional layers that simulate the performance of the Frangi filter and facilitate gradients during training. This compact convolutional network is optimized with raw images as input and filtered outputs as the target[24].

Additionally, we incorporate a noise-augmentation training strategy to reduce noise for low-dose CT and improve the robustness of HiPaS to CT noise. Specifically, we add simulated noise onto the low-resolution CT slices to get the noisy slices $noise \circ \mathrm{slice}^{lr}$ (Supplementary Figure 6.b), expecting I2SR to recover the raw CT slices from these noisy inputs. The simulated noise follows the Poisson decay law[57]. The total objective function for I2SR is thus formulated as

$$\mathcal{L}^{I2SR} = \| \mathrm{slice}^{\mathrm{re}} - \mathrm{slice}^{\mathrm{gt}} \|^2 + \| O_F(\mathrm{slice}^{\mathrm{re}}) - O_F(\mathrm{slice}^{\mathrm{gt}}) \|^2,$$

where

$$\mathrm{slice}^{\mathrm{re}} = \mathrm{I2SR}(\mathrm{noise} \circ \mathrm{slice}^{\mathrm{lr}}).$$

### *Artery-vein segmentation with saliency-transmission segmentation module*

Current algorithms for pulmonary artery-vein segmentation are largely hindered by three primary factors[19]: the substantial morphological disparities between extra- and intra-pulmonary blood vessels, the limited representation of intra-pulmonary vessels in the input CT scan, and the long-range context correlation to distinguish arteries and veins. Due to the difficulty in simultaneously learning and distinguishing the distinct morphological features of the extra- and intra-pulmonary blood vessels, a single segmentation model may be insufficient in achieving satisfactory segmentation results. The complex morphology of blood vessels can also exacerbate topology issues, leading to problems such as omission and discontinuity of segmentation results. Additionally, local information alone is often insufficient to distinguish between arteries and veins, which requires long-range contextual information instead, even from the hilum. Otherwise, misclassification will easily exist for intrapulmonary artery-vein segmentation.

In response, we have developed a novel Saliency-Transmission Segmentation (STS) module that effectively addresses all of the aforementioned concerns (Fig 1.c, right). The module design is inspired by the manual segmentation process, in which experts first identify the low-level arteries and veins, and then infer higher-level vessels based on spatial continuity, topological structure, and physiological information. Following such an idea, we implement a deep learning approach that conducts segmentation in a hierarchical manner, with lower-level results providing contextual priors to guide higher-level segmentation. This aligns with the step-wise process of manual labeling. Meanwhile, the network in each stage can better focus on specific vessel topology instead of simultaneously handling vessels with various topologies. Specifically, the module contains an innovative prior-map transmission block that passes probability maps between layers based on preliminary vessel delineation (Supplementary Figure 5.b). This transmission block converts the segmentation probability map of the lower-level vessel branches from the previous networks into spatial weights, which are then applied to the next segmentation.

We first divide the arteries and veins of the annotations into four levels, denoted as

$$[A^0,V^0],[A^1,V^1],[A^2,V^2],[A^3,V^3].$$

Here $A$ is the artery annotations and $V$ is the vein annotations. $[A^0,V^0]$ represents the cardinal arteries and veins inside the heart; $[A^1,V^1]$ includes the arteries and veins located as the hilum with 1 or 2 levels of vessel branches; $[A^2,V^2]$ includes the arteries and veins with 3 to 5 levels of vessel branches, and $[A^3,V^3]$ includes all the left visible vessels inside the lung. Supplementary Figure 6.c provides two examples about the divide of the arteries and veins. The branch levels are obtained by first extracting the vessel skeleton[58], and then transforming the skeleton into a vessel tree[59].

Without loss of generality, we consider a resampled CT scan $Y$ as the input, and the goal is to achieve a binary segmentation mask with the same dimensionality. For the artery-vein segmentation of the $i$-th level $[A^i,V^i]$ $(i\geqslant 1)$, assuming that we have obtained the possibility map of the $i-1$ level segmentation as $[P_{\mathrm{A}}^{i-1},P_{\mathrm{V}}^{i-1}]$, the prior-map transmission will incorporate the possibility map with the input CT scans to generate an updated input data, denoted as

$$\hat{Y}^i = Y \oplus C_{\mathrm{T}}(P_{\mathrm{A}}^{i-1},P_{\mathrm{V}}^{i-1}).$$

Here $C_{\mathrm{T}}$ is a size-preserved convolution and $\oplus$ represents the concatenation operation along the channel dimension. In practice, we use two layers of $3\times3\times3$ convolutional layers with residual connection as $C_{\mathrm{T}}$. Then the segmentation process becomes:

$$[P_{\mathrm{A}}^i,P_{\mathrm{V}}^i] = S^i\left(\hat{Y}^i;\theta^i\right).$$

$\theta^i$ is the weight for the segmentation network $S^i$, and the network architecture is shown in Supplementary Figure 5.c and d. For the first-level segmentation $[A^0,V^0]$, its previous segmentation map will be replaced by the all-zero matrix. To further enhance the segmentation performance, we replace the input $Y$ with a concatenated representation of the CT scan and its refined vascular features, a process discussed earlier,

$$Y \oplus O_F(Y) \rightarrow Y.$$

Based on branch-level labeling, we design a weighted dice-loss which can automatically balance weights for different blood-vessel levels. The weighted dice loss is defined as follows

$$\mathcal{L}_{\mathrm{DSC}} = -\left(\frac{P^0\cdot T^0}{P^0+T^0} + \sum_{i=1}^{3}\left(w^i\frac{\Delta P^i\cdot\Delta T^i}{\Delta P^i+\Delta T^i}\right)\right).$$

Here $P$ and $T$ represent the predicted segmentation results and the ground-truth segmentation, while $\Delta$ means the differences between $i$ and $i-1$ levels of the vessel tree.

$$\begin{cases} \Delta P^{i} = P^{i} - P^{i-1} \\ \Delta T^{i} = T^{i} - T^{i-1} \end{cases}$$

The weight of low-level branches $w^{i}$ should be lower than the weight of high-level branches to increase the attention to the intra-pulmonary high-level branches in the training process, and we set $w^{i}$ to be inversely proportional to the volume of $\Delta T^{i}$

$$w^{i} = \frac{V(T^{0})}{V(\Delta T^{i})}.$$

Here $V(\cdot)$ means the calculation of volumes, and we use the counts of voxel numbers as the volumes in our study. The weighted dice loss is respectively computed for arteries and veins. To prevent misclassification between arteries and veins, we also incorporate overlap-cross loss in our methodology:

$$\mathcal{L}_{\text{overlap}} = \frac{P_A \cdot T_V + P_V \cdot T_A}{P + T}.$$

In conclusion, the overall loss function is

$$\mathcal{L} = \mathcal{L}_{\text{DSC}} + \mathcal{L}_{\text{overlap}}.$$

***Transfer learning from CTPA to non-contrast CT***

To address the challenge of directly segmenting arteries and veins in non-contrast CT, which presents difficulties for both radiologists and networks due to low imaging contrast, we introduce a novel transfer learning approach. Initially, radiologists annotated arteries and veins on all the CTPA scans (n=315). Our network was trained on these annotated scans to create an initial model. We then developed a transformation model, utilizing methods from [62], to generate synthetic non-contrast CT images from the corresponding CTPA scans while preserving physiological information. This transformation model reduced the imaged blood vessel intensities (from over 200 HU to approximately 50 HU) to resemble non-enhanced vessels while maintaining the signals of surrounding tissues. The network was subsequently fine-tuned on the transformed non-contrast CT images and then applied to segment arteries and veins on native non-contrast CT scans. Radiologists used these initial segmentation predictions as a basis for manually labeling the arteries and veins.

***Paired non-contrast CT and CTPA with DSCTPA***

Fourteen patients with suspected pulmonary embolism underwent digital subtraction computed tomography pulmonary angiography (DSCTPA)[29] utilizing a GE CT scanner (Revolution GSI). Patients were positioned supine, with full lung coverage from the apex through the diaphragm achieved using the following parameters: the tube voltage equaling 120 kV, X-ray tube current of 330 mA, and 1.00 mm slice thickness. The CT scans were reconstructed with the filtered back-projection[63] reconstruction algorithm. Prior to scanning, patients underwent respiratory training with the scanner to simulate breathing patterns required during image acquisition. Patients were coached to breathe as consistently as possible throughout the examination. Non-contrast CT was performed first, followed immediately by CTPA within a 30-second interval. For CTPA, iodixanol contrast (Visipaque 320, GE Healthcare, Shanghai) was administered at 30-40 mL through a CT-specific high-pressure syringe at 3.5-5 mL/s flow rate. A 40 mL saline flush followed contrast administration. The non-contrast CT scans were utilized to compare the segmentation performance of HiPaS against CTPA. Arteries and veins were first manually segmented on non-contrast and CTPA scans, respectively; then we computed DSC between the segmentation results achieved by HiPaS and the manual annotation.

***Semi-automatic artery-vein segmentation***

Here we describe the method of semi-automatic artery-vein segmentation, which is the prevalent clinical artery-vein segmentation method in the absence of HiPaS, requiring the expertise of experienced radiologists for its execution. The semi-automatic annotation process commences with the radiologist uploading a CT scan into the CT visualization software. The software employs a user-friendly interface where radiologists can navigate through the scan slices. Once the CT scan is loaded, the radiologist initiates the artery-vein segmentation process by denoting the approximate location and morphology of cardinal arteries and veins inside the heart. The algorithm will then automatically derive the segmentations for the cardinal arteries and veins via our in-house algorithms under the guidance of initial annotations given by radiologists. Here the radiologist will review the results and interactive editing the segmentation results. Then this annotation will act as a seed point from which the algorithms extrapolate to identify and segment the connected vessels. The algorithms used may include region-growing techniques, thresholding, or edge detection methods that utilize the inherent contrast between the blood vessels and the surrounding lung parenchyma. Finally, the software will derive a segmentation result including both vessel trunks and intrapulmonary arteries and veins. Such semi-automated segmentation requires approximately thirty minutes per CT scan.

***Dataset summary***

This section provides a comprehensive overview of the datasets used in the development and evaluation of HiPaS, as well as in the anatomical study, to facilitate data verification for readers. Initially, we utilized 17,817 CT volumes from two sources: 13,000 from [60] and 4,817 from [61], for model pretraining. Subsequently, we established a dataset of 1,073 volumes with detailed artery-vein segmentation annotations, comprising 315 CTPA and 758 non-contrast CT volumes. During the annotation process, a "human-in-the-loop" strategy was employed. Radiologists annotated the 315 CTPA volumes first. HiPaS was then trained on these data and applied to non-contrast CT volumes for initial segmentation. Radiologists reviewed and revised these annotations, obtaining the final annotations for all 1,073 volumes. For the formal training of HiPaS, we used 875 CT volumes (all 315 CTPA and 560 non-contrast) and tested the model on a set of 198 non-contrast CT volumes (142 normal-resolution and 56 low-resolution). An experiment involving 14 DSCTPA cases was conducted to demonstrate the non-inferiority of HiPaS segmentation on non-contrast CT compared to CTPA. Additionally, 50 CT volumes were used to qualitatively assess the clinical utility of HiPaS segmentation. Finally, for the anatomical study, we collected a large-scale dataset of 11,784 CT volumes from six sites, including the manually annotated 1,073 volumes and an additional 10,711 volumes.

***Experimental settings***

Direct training of neural networks for the segmentation of arteries and veins in non-contrast CT images presents challenges. Therefore, we first train the network on CTPA data, utilizing the parameters for initialization of the model for non-contrast CT segmentation. The voxel intensity of all scans is truncated within the Hounsfield Unit (HU) window of [-1000, 600] and normalization to [0, 1]. Due to the GPU memory limit, CT scans are cropped into sub-volume cubes of the size $192 \times 192 \times 128$ for segmentation tasks. We use PyTorch to implement the proposed method. The Adam optimizer is used for the segmentation network with an initial learning rate of $1 \times 10^{-4}$. The decay of the first-order momentum gradient is 0.9, and the decay of the second-order momentum gradient is 0.999. In the experiments, model training is executed on a Linux workstation with NVIDIA RTX A100. Current parameters perform well for our tasks but are not necessarily optimum. Adjustments may be conducted for specific tasks.

***Statistical analyses***

To conduct a comprehensive quantitative assessment of the segmentation performance, we employ various evaluation metrics, including dice similarity coefficient (DSC), sensitivity (SEN), specify (SPE) , misclassification Score (MCS), vessel branch counts (BC), vessel skeleton length (SL), and the 95% Hausdorff Distance (HD95). The calculation of DSC, SEN, SPE, and HD95 adhere to the conventional definitions. The MCS is employed to gauge the misclassification rate between arteries and veins, which is defined as $\mathrm{MCS} = \frac{P_A \cdot T_V + P_V \cdot T_A}{P + T}$. The metrics of branch counts and skeleton length are utilized to evaluate the abundance of segmented branches, providing valuable insights into the distinctive characteristics of pulmonary arteries and veins. To facilitate the computations, we first re-sample all segmentation results to a standardized space with a spatial resolution of $\frac{334}{512} \times \frac{334}{512} \times 1.00\ \mathrm{mm}^3$. Subsequently, we employ the algorithm developed by [58] to automatically extract the vessel skeletons by reconstructing an octree data structure. The skeleton length is determined by counting the number of pixels within the extracted vessel skeletons, while the branch counts are obtained by counting the number of bifurcations. HD95 is employed to quantify the dissimilarities between the boundaries of the segmentation results and the annotated data. Smaller values of HD95 always indicate superior segmentation outcomes. The calculation of lung volume is performed on the segmented lung from CT scans, following the method proposed in [34]. The volume of the segmented lung subtracted the volume of all intrapulmonary blood vessels is defined as the lung volume in our study.

Data are reported as $\mathrm{mean} \pm \mathrm{std}$ unless stated otherwise. In some cases, where arterial and vein indicators need to be described separately, we use the form $\mathrm{mean}_A \pm \mathrm{std}_A / \mathrm{mean}_V \pm \mathrm{std}_V$ for arteries and veins, respectively. The Wilcoxon signed-rank test is performed to evaluate the distribution of two paired groups (such as DSC and SEN achieved by the two methods), otherwise, the Mann-Whitney U test is used. The association between vessel abundance and sex, age, and lung volume is validated with both multiple linear regression and the Chi-square test. $p < 0.05$ is considered statistically significant in this study. $P$ values are specified in the figures and tables as *$p$<0.05, **$p$<0.01, ***$p$<0.001, ****$p$<0.0001, NS, not significant. All statistical analyses are performed in Python 3.8 and R version 4.3.0.

**Ethics statement**

The patient data were collected from The First Affiliated Hospital of Harbin Medical University, the Fourth Affiliated Hospital of Harbin Medical University, Mudanjiang First People's Hospital, China-Japan Friendship Hospital, Shanghai Renji Hospital, and Guangdong Provincial People's Hospital, following the approval from the Institutional Review Board. The experiment using DSCTPA was approved by the Fourth Affiliated Hospital of Harbin Medical University. The study was also approved by the Institutional Biosafety and Bioethics Committee at King Abdullah University of Science and Technology. Informed consent was waived in the training cohort and the inpatient cohort due to the retrospective nature of the study.

**Code Availability**

The code is publicly available under https://github.com/Arturia-Pendragon-Iris/HiPaS_AV_Segmentation.

**Data Availability**

Sample data with annotations are given at https://github.com/Arturia-Pendragon-Iris/HiPaS_AV_Segmentation. The remaining datasets used in this study are currently not permitted for public release by the respective institutional review boards. All data provided are anonymized to protect the privacy of the patients who participated in the studies, in line with applicable laws and regulations. The anatomical study data can be provided based on a written data request, pending scientific review, and writer cooperation agreement.

**Competing interests**

The authors declare no competing interests.

**Acknowledgment**

This publication is based upon work supported by the King Abdullah University of Science and Technology (KAUST) Office of Research Administration (ORA) under Award No URF/1/4352-01-01, FCC/1/1976-44-01, FCC/1/1976-45-01, REI/1/5234-01-01, REI/1/5414-01-01, REI/1/5289-01-01, REI/1/5404-01-01.

**References**

[1] Nabel E G. Cardiovascular disease[J]. New England Journal of Medicine, 2003, 349(1): 60-72.

[2] Tarride J E, Lim M, DesMeules M, et al. A review of the cost of cardiovascular disease[J]. Canadian Journal of Cardiology, 2009, 25(6): e195-e202.

[3] Ye Y, Sing C W, Hubbard R, et al. Prevalence, incidence, and survival analysis of interstitial lung diseases in Hong Kong: a 16-year population-based cohort study[J]. The Lancet Regional Health–Western Pacific, 2024, 42.

[4] Goldhaber S Z, Morrison R B. Pulmonary embolism and deep vein thrombosis[J]. Circulation, 2002, 106(12): 1436-1438.

[5] McLaughlin V V, McGoon M D. Pulmonary arterial hypertension[J]. Circulation, 2006, 114(13): 1417-1431.

[6] Sadeghi A H, Maat A P W M, Taverne Y J H J, et al. Virtual reality and artificial intelligence for 3-dimensional planning of lung segmentectomies[J]. JTCVS techniques, 2021, 7: 309-321.

[7] Wang D, Pan Y, Durumeric O C, et al. PLOSL: Population learning followed by one shot learning pulmonary image registration using tissue volume preserving and vesselness constraints[J]. Medical image analysis, 2022, 79: 102434.

[8] Pu J, Gezer N S, Ren S, et al. Automated detection and segmentation of pulmonary embolisms on computed tomography pulmonary angiography (CTPA) using deep learning but without manual outlining[J]. Medical Image Analysis, 2023, 89: 102882.

[9] Shen Z, Feydy J, Liu P, et al. Accurate point cloud registration with robust optimal transport[J]. Advances in Neural Information Processing Systems, 2021, 34: 5373-5389.

[10] Hulten E A, Carbonaro S, Petrillo S P, et al. Prognostic value of cardiac computed tomography angiography: a systematic review and meta-analysis[J]. Journal of the American College of Cardiology, 2011, 57(10): 1237-1247.

[11] Koelemay M J W, Nederkoorn P J, Reitsma J B, et al. Systematic review of computed tomographic angiography for assessment of carotid artery disease[J]. Stroke, 2004, 35(10): 2306-2312.

[12] Pasternak J J, Williamson E E. Clinical pharmacology, uses, and adverse reactions of iodinated contrast agents: a primer for the non-radiologist[C]//Mayo Clinic Proceedings. Elsevier, 2012, 87(4): 390-402.

[13] Singh J, Daftary A. Iodinated contrast media and their adverse reactions[J]. Journal of nuclear medicine technology, 2008, 36(2): 69-74.

[14] Jean ‐ Marc I, Emmanuelle P, Philippe P, et al. Allergy ‐ like reactions to iodinated contrast agents. A critical analysis[J]. Fundamental & clinical pharmacology, 2005, 19(3): 263-281.

[15] Nguyen T N, Abdalkader M, Nagel S, et al. Noncontrast computed tomography vs computed tomography perfusion or magnetic resonance imaging selection in late presentation of stroke with large-vessel occlusion[J]. JAMA neurology, 2022, 79(1): 22-31.

[16] Buzug T M. Computed tomography[M]//Springer handbook of medical technology. Berlin, Heidelberg: Springer Berlin Heidelberg, 2011: 311-342.

[17] Nardelli P, Jimenez-Carretero D, Bermejo-Pelaez D, et al. Pulmonary artery–vein classification in CT images using deep learning[J]. IEEE transactions on medical imaging, 2018, 37(11): 2428-2440.

[18] Qin Y, Zheng H, Gu Y, et al. Learning tubule-sensitive CNNs for pulmonary airway and artery-vein segmentation in CT[J]. IEEE transactions on medical imaging, 2021, 40(6): 1603-1617.

[19] Pan L, Li Z, Shen Z, et al. Learning multi-view and centerline topology connectivity information for pulmonary artery–vein separation[J]. Computers in Biology and Medicine, 2023, 155: 106669.

[20] Wu Y, Qi S, Wang M, et al. Transformer-based 3D U-Net for pulmonary vessel segmentation and artery-vein separation from CT images[J]. Medical & Biological Engineering & Computing, 2023, 61(10): 2649-2663.

[21] Voelkel N F, Tuder R M. Hypoxia-induced pulmonary vascular remodeling: a model for what human disease?[J]. The Journal of clinical investigation, 2000, 106(6): 733-738.

[22] Chu Y, Zhang Y, Han Z, et al. Improving Representation of High-frequency Components for Medical Foundation Models[J]. arXiv preprint arXiv:2407.14651, 2024.

[23] Wang G, Ye J C, De Man B. Deep learning for tomographic image reconstruction[J]. Nature Machine Intelligence, 2020, 2(12): 737-748.

[24] Chu Y, Zhou L, Luo G, et al. Topology-Preserving Computed Tomography Super-Resolution Based on Dual-Stream Diffusion Model[C]//International Conference on Medical Image Computing and Computer-Assisted Intervention. Cham: Springer Nature Switzerland, 2023: 260-270.

[25] Shan H, Padole A, Homayounieh F, et al. Competitive performance of a modularized deep neural network compared to commercial algorithms for low-dose CT image reconstruction[J]. Nature Machine Intelligence, 2019, 1(6): 269-276.

[26] Liu J, Jiang H, Ning F, et al. DFSNE-Net: Deviant feature sensitive noise estimate network for low-dose CT denoising[J]. Computers in Biology and Medicine, 2022, 149: 106061.

[27] Brady S L, Trout A T, Somasundaram E, et al. Improving image quality and reducing radiation dose for pediatric CT by using deep learning reconstruction[J]. Radiology, 2021, 298(1): 180-188.

[28] Dong Z, He Y, Qi X, et al. MNet: Rethinking 2D/3D Networks for Anisotropic Medical Image Segmentation[C]//Thirty-First International Joint Conference on Artificial Intelligence {IJCAI-22}. International Joint Conferences on Artificial Intelligence Organization, 2022: 870-876.

[29] Winer-Muram H T, Rydberg J, Johnson M S, et al. Suspected acute pulmonary embolism: evaluation with multi–detector row CT versus digital subtraction pulmonary arteriography[J]. Radiology, 2004, 233(3): 806-815.

[30] Zhang M, Wu Y, Zhang H, et al. Multi-site, multi-domain airway tree modeling[J]. Medical Image Analysis, 2023, 90: 102957.

[31] Eisma J J, McKnight C D, Hett K, et al. Deep learning segmentation of the choroid plexus from structural magnetic resonance imaging (MRI): validation and normative ranges across the adult lifespan[J]. Fluids and Barriers of the CNS, 2024, 21(1): 1-13.

[32] Wolny A, Cerrone L, Vijayan A, et al. Accurate and versatile 3D segmentation of plant tissues at cellular resolution[J]. Elife, 2020, 9: e57613.

[33] Isensee F, Jaeger P F, Kohl S A A, et al. nnU-Net: a self-configuring method for deep learning-based biomedical image segmentation[J]. Nature methods, 2021, 18(2): 203-211.

[34] Zhou L, Meng X, Huang Y, et al. An interpretable deep learning workflow for discovering subvisual abnormalities in CT scans of COVID-19 inpatients and survivors[J]. Nature Machine Intelligence, 2022, 4(5): 494-503.

[35] Xu X, Wang B, Ren C, et al. Age-related impairment of vascular structure and functions[J]. Aging and disease, 2017, 8(5): 590.

[36] Ji H, Kwan A C, Chen M T, et al. Sex differences in myocardial and vascular aging[J]. Circulation Research, 2022, 130(4): 566-577.

[37] Shahin Y, Alabed S, Alkhanfar D, et al. Quantitative CT evaluation of small pulmonary vessels has functional and prognostic value in pulmonary hypertension[J]. Radiology, 2022, 305(2): 431-440.

[38] Hueper K, Vogel-Claussen J, Parikh M A, et al. Pulmonary microvascular blood flow in mild chronic obstructive pulmonary disease and emphysema. The MESA COPD Study[J]. American journal of respiratory and critical care medicine, 2015, 192(5): 570-580.

[39] Karnati S, Seimetz M, Kleefeldt F, et al. Chronic obstructive pulmonary disease and the cardiovascular system: vascular repair and regeneration as a therapeutic target[J]. Frontiers in Cardiovascular Medicine, 2021, 8: 649512.

[40] Blanco I, Piccari L, Barberà J A. Pulmonary vasculature in COPD: the silent component[J]. Respirology, 2016, 21(6): 984-994.

[41] Molgat-Seon Y, Peters C M, Sheel A W. Sex-differences in the human respiratory system and their impact on resting pulmonary function and the integrative response to exercise[J]. Current Opinion in Physiology, 2018, 6: 21-27.

[42] Morris H, Denver N, Gaw R, et al. Sex differences in pulmonary hypertension[J]. Clinics in Chest Medicine, 2021, 42(1): 217-228.

[43] Gillis E E, Sullivan J C. Sex differences in hypertension: recent advances[J]. Hypertension, 2016, 68(6): 1322-1327.

[44] Jarman A F, Mumma B E, Singh K S, et al. Crucial considerations: Sex differences in the epidemiology, diagnosis, treatment, and outcomes of acute pulmonary embolism in non‐pregnant adult patients[J]. Journal of the American College of Emergency Physicians Open, 2021, 2(1): e12378.

[45] Sedhom R, Megaly M, Elbadawi A, et al. Sex differences in management and outcomes among patients with high-risk pulmonary embolism: a nationwide analysis[C]//Mayo Clinic Proceedings. Elsevier, 2022, 97(10): 1872-1882.

[46] Gazdar A F, Thun M J. Lung cancer, smoke exposure, and sex[J]. Journal of Clinical Oncology, 2007, 25(5): 469-471.

[47] Tai R, Dunne R M, Trotman-Dickenson B, et al. Frequency and severity of pulmonary hemorrhage in patients undergoing percutaneous CT-guided transthoracic lung biopsy: single-institution experience of 1175 cases[J]. Radiology, 2016, 279(1): 287-296.

[48] Olson T P, Beck K C, Johnson B D. Pulmonary function changes associated with cardiomegaly in chronic heart failure[J]. Journal of cardiac failure, 2007, 13(2): 100-107.

[49] Disthabanchong S, Boongird S. Role of different imaging modalities of vascular calcification in predicting outcomes in chronic kidney disease[J]. World journal of nephrology, 2017, 6(3): 100.

[50] Sakao S, Voelkel N F, Tatsumi K. The vascular bed in COPD: pulmonary hypertension and pulmonary vascular alterations[J]. European respiratory review, 2014, 23(133): 350-355.

[51] Brandts A, Westenberg J J M, Van Elderen S G C, et al. Site-specific coupling between vascular wall thickness and function: an observational MRI study of vessel wall thickening and stiffening in hypertension[J]. Investigative radiology, 2013, 48(2): 86-91.

[52] He K, Chen X, Xie S, et al. Masked autoencoders are scalable vision learners[C]//Proceedings of the IEEE/CVF conference on computer vision and pattern recognition. 2022: 16000-16009.

[53] Tian K, Jiang Y, Lin C, et al. Designing BERT for Convolutional Networks: Sparse and Hierarchical Masked Modeling[C]//The Eleventh International Conference on Learning Representations. 2022.

[54] Chen H, Zhang Y, Kalra M K, et al. Low-dose CT with a residual encoder-decoder convolutional neural network[J]. IEEE transactions on medical imaging, 2017, 36(12): 2524-2535.

[55] Frangi A F, Niessen W J, Vincken K L, et al. Multiscale vessel enhancement filtering[C]//Medical Image Computing and Computer-Assisted Intervention—MICCAI'98: First International Conference Cambridge, MA, USA, October 11–13, 1998 Proceedings 1. Springer Berlin Heidelberg, 1998: 130-137.

[56] Hachaj T, Piekarczyk M. High-Level Hessian-Based Image Processing with the Frangi Neuron[J]. Electronics, 2023, 12(19): 4159.

[57] Leuschner J, Schmidt M, Baguer D O, et al. The lodopab-ct dataset: A benchmark dataset for low-dose ct reconstruction methods[J]. arXiv preprint arXiv:1910.01113, 2019.

[58] Cornea N D, Silver D, Yuan X, et al. Computing hierarchical curve-skeletons of 3D objects[J]. The Visual Computer, 2005, 21: 945-955.

[59] Yu W, Zheng H, Gu Y, et al. Tnn: Tree neural network for airway anatomical labeling[J]. IEEE Transactions on Medical Imaging, 2022, 42(1): 103-118.

[60] Draelos R L, Dov D, Mazurowski M A, et al. Machine-learning-based multiple abnormality prediction with large-scale chest computed tomography volumes[J]. Medical image analysis, 2021, 67: 101857.

[61] Colak E, Kitamura F C, Hobbs S B, et al. The RSNA pulmonary embolism CT dataset[J]. Radiology: Artificial Intelligence, 2021, 3(2): e200254.

[62] Gu X, Liu Z, Zhou J, et al. Contrast-enhanced to non contrast CT transformation via an adjacency content-transfer-based deep subtraction residual neural network[J]. Physics in Medicine & Biology, 2021, 66(14): 145017.

[63] Willemink M J, Noël P B. The evolution of image reconstruction for CT—from filtered back projection to artificial intelligence[J]. European radiology, 2019, 29: 2185-2195.